\documentclass[letterpaper, 10 pt, conference]{ieeeconf}  

\IEEEoverridecommandlockouts                              

\usepackage{cite}
\usepackage{amsmath,amssymb,amsfonts}
\usepackage{algorithmic}
\usepackage{graphicx}
\usepackage{textcomp}
\usepackage{xcolor}
\usepackage{bm}
\def\BibTeX{{\rm B\kern-.05em{\sc i\kern-.025em b}\kern-.08em
    T\kern-.1667em\lower.7ex\hbox{E}\kern-.125emX}}
\def\awano{}
\def\ohara{}

\title{\LARGE \bf
Binary Neural Network in Robotic Manipulation: Flexible Object Manipulation for Humanoid Robot Using Partially Binarized Auto-Encoder on FPGA
}

\author{Satoshi Ohara$^{1}$, Tetsuya Ogata$^{2}$, and Hiromitsu Awano$^{3}$
\thanks{*This work was supported by JST, PRESTO Grant No. JP-MJPR18M1.}
\thanks{$^{1}$Satoshi Ohara is with the Department of Information Systems Engineering, Graduate School of Information Science and Technology, Osaka University, 1-5 Yamadaoka, Suita, Osaka 565-0871, JAPAN
        {\tt\small s-ohara@ist.osaka-u.ac.jp}}%
\thanks{$^{2}$Tetsuya Ogata is with the Department of Intermedia Art and Science, School of Fundamental Science and Engineering, Waseda University, 3-4-1 Ohkubo, Shinjuku, Tokyo, 169-8555, JAPAN
        {\tt\small ogata@waseda.jp}}%
\thanks{$^{3}$Hiromitsu Awano is with the Department of Communication and Computer Engineering, Graduate School of Infomatics, Kyoto University, Yoshida-hon-machi, Sakyo, Kyoto 606-8501, JAPAN
        {\tt\small awano@i.kyoto-u.ac.jp}}%
}

\begin{document}

\maketitle
\thispagestyle{empty}
\pagestyle{empty}

\begin{abstract}
\label{sec:abstract}
A neural network based flexible object manipulation system for a humanoid robot on FPGA is proposed. Although the manipulations of flexible objects using robots attract ever increasing attention since these tasks are the basic and essential activities in our daily life, it has been put into practice only recently with the help of deep neural networks. However such systems have relied on GPU accelerators, which cannot be implemented into the space limited robotic body. Although field programmable gate arrays (FPGAs) are known to be energy efficient and suitable for embedded systems, the model size should be drastically reduced since FPGAs have limited on-chip memory. To this end, we propose ``partially'' binarized  deep convolutional auto-encoder technique, where only an encoder part is binarized to compress model size without degrading the inference accuracy. 
The model implemented on Xilinx ZCU102 achieves 41.1 frames per second with a power consumption of 3.1~W, {\awano{which corresponds to 10$\times$ and 3.7$\times$ improvements from the systems implemented on Core i7 6700K and RTX 2080 Ti, respectively.}}
\end{abstract}

\section{Introduction}
\label{sec:intro}
{\awano{Robotic manipulation of deformable objects such as clothes attracts ever increasing attention since these tasks are one of the basic and essential activities in our daily life.}}
This type of manipulation is one of the most challenging {\awano{tasks since the shape of flexible objects changes largely for different end-effector trajectories. To solve this problem, applications of end-to-end learning, where robots learn directly from human demonstrations, have been reported~\ohara{\cite{kawaharazuka2019dynamic, kase2018put, saito2019real}.}}}
{\awano{One of the state-of-the-art method proposed by}} Pin-Chu Yang et al. {\awano{introduced a two-stage deep learning model where a deep convolutional autoencoder (DCAE) extracts low-dimensional image features from raw camera inputs, followed by a fully connected deep time delay neural network for predicting the temporal sequences of the image features and the joint angles of the robot. During the inference phase, the raw image captured by {\ohara{the}} camera mounted on robot's head and current joint angles of the robot are provided to the trained model which, in turn, outputs the joint angles at the next time step~\cite{yang2016repeatable}. Although they succeeded in folding a towel, their approach is implemented on a GPU-based platform consuming hundreds of watts, thus cannot be embedded into the space-limited robotic body.}}
{\awano{To alleviate this problem, we propose an FPGA-based system for robotic manipulation of flexible objects. Contrary to instruction-based processors like CPU or GPU, FPGAs are configured by hardware circuit specifications, which drastically improves the energy efficiency. However, FPGA platform{\ohara{s}} usually has less memory space compared to GPU platform{\ohara{s}}, and hence an aggressive model size compression is definitely required. One of the well-known model compression method is ``binarized'' neural networks (BNNs), where the synaptic weights and network activations are approximately represented by 1-bit values~\cite{courbariaux2016binarized}. Surprisingly, even when the networks are binarized, the BNNs have acceptable performance on several tasks such as MNIST or CIFAR10 image classifications.

In spite of the successful applications of BNNs in the image classification tasks, we find that the naive binarization of DCAE results in the corruption of the extracted image features since the binarization blocks gradients to well propagate through the network.
To alleviate this, we propose ``partial'' binarization method named {\it partially}-binarized DCAE (PB-DCAE), where only synaptic weights and activations in the encoder part are binarized whereas those in the decoder part are left unchanged. The full precision decoder part help gradient to propagate through the deep layers, resulting in the well self-organized image features. Note here that the decoder part is only required during training and is removed for inference. Hence, our method does not incur any additional memory space compared to the fully binarized DCAE.
}}
This paper makes the following contributions:
\begin{itemize}
    \item {\awano{To the best of our knowledge, this is the first study reporting the successful application of BNN into robotic domain.}}
    \item {\awano{A}} lightweight neural network model {\awano{named PB-DCAE}} for end-to-end learning of flexible object manipulation by a humanoid robot {\awano{is proposed}}.
    \item {\awano{The proposed model implemented on Xilix ZCU102 board demonstrates that 10$\times$ and 3.7$\times$ more improvement in energy efficiency compared to that implemented on Core i7 6700K CPU {\ohara{and}} RTX2080Ti GPU, respectively.}}
\end{itemize}
\section{PRELIMINARIES}
\label{sec:preliminaries}
\subsection{{\awano{End-to-End Learning of Robotic Object Manipulation}}}

{\awano{End-to-end learning of robotic manipulation is composed of three phases: (1) data collection, (2) model training, and (3) model deployment on a robot.

\subsubsection{Data collection}
A robot teleoperating environment is constructed, where a human instructor can control robot end-effectors via 3D spatial input devices. During the teleoperation, a robot is operated by a human instructor while recording the states and the corresponding actions from its own sensors, e.g., a camera mounted on the robot's head to capture changes in the physical environment and angle sensors to monitor the movements of the robot arm{\ohara{s}}. Note here that since the state and actions are recorded by robot's own sensors, no coordinate transformations are required to train the model.

\subsubsection{Model training}
The typical network architecture to enable end-to-end robotic object manipulation is to combine two different models: a DCAE to extract low-dimensional features from raw images and a recurrent neural network (RNN) to model temporal sequence{\ohara{s}} of image features and joint angles~\cite{yang2016repeatable, kase2019learning}. DCAE is a symmetrical hourglass-shaped convolutional neural network composed of an ``encoder'' and a ``decoder,'' which is used for {\ohara{dimensionality}} reduction~\cite{masci2011stacked}. The training objective of the DCAE is to reconstruct the original images provided to the encoder. Since its middle layers have fewer neurons compared to the input/output layers, the trained DCAE can create a ``compressed'' representation of the original image in its middle layer. After the training, the decoder is removed and the output of {\ohara{the}} encoder is directed to the RNN for joint angle prediction.
}}


{\awano{RNN is a class of neural network{\ohara{s}} where the outputs are fed as input recursively, which enable the RNN to have internal memory and to behave temporally.}}
RNN can process sequential information and maintain robustness against instantaneous noise. Therefore, RNN are well suited for robot manipulation and have been used for various tasks such as drawing\cite{sasaki2015neural}, tool usage\cite{takahashi2015tool}, and "put-in-box" tasks\cite{kase2018put}.
Given the image features extracted by the encoder and the current joint angles of {\ohara{the}} robot, the RNN is trained to successfully predict the joint angles at the next time step.

{\awano{
\subsubsection{Model deployment}
Firstly, the camera image is fed into the trained encoder to extract the low-dimensional feature. Secondly, the feature vector is concatenated with the current joint angles, which is provided to the RNN for predicting the joint angles at {\ohara{the}} next time step. Finally, the predicted joint angles are transmitted to {\ohara{the}} robot controller to actually rotate joints. The above steps are repeated until {\ohara{the}} robot complete object manipulation.
}}

\subsection{{\awano{Binarized Neural Networks}}}

{\awano{To reduce the model size, \cite{courbariaux2016binarized} has proposed binarization of synaptic parameters and activations, i.e., a real-valued weight or activation
is approximated
by
either $+1$ or $-1$.
The real-valued input tensor is binarized using the following activation function:
\begin{eqnarray}
  \label{eq:kettei}
  x^b = Sign(x) = \begin{cases}
  +1 & x \geq 0,\\
  -1 & otherwise.
  \end{cases}
\end{eqnarray}
However, since the derivative of the sign function is almost zero everywhere, naive binarization leads to poor network performance. To alleviate this, \cite{courbariaux2016binarized} proposed a technique called ``straight-through estimator,'' where the derivative of the sign function is substituted by that of hard hyperbolic tangent, i.e.,
\begin{align}
    \frac{ \partial }{ \partial x } Sign(x) & \approx \begin{cases}
    1 & \left| x \right| < 1,\\
    0  & otherwise.
    \end{cases}
\end{align}
By constraining the weights and activations to be binary, we can (1) dramatically reduce model size, e.g., considering float32 as baseline, binarization achieves 1/32 model size reduction, and (2) multiply-and-accumulate (MAC) operations can be realized by using only XNOR gates and 1's counters.
}}


\subsection{Batch Normalization}

Even with ``straight-through estimator,'' training of BNN is unstable and gradients tend to explode frequently. Hence, BNN usually comes with batch normalization (BN) layer which basically standardizes activations, i.e., it re-scales and re-centers input activations so that their means and variances become zero and one, respectively\cite{ioffe2015batch}. Let $x$ is the input to the BN layer, it computes:
\begin{align}
    y &= \gamma \frac{ x - \mu }{\sqrt{\sigma^2 + \epsilon}}, \label{eq:bn}
\end{align}
where $y$ is the output of the BN layer, $\gamma$ and $\beta$ are trainable parameters, $\mu$ and $\sigma^2$ are the empirical mean and variances of mini-batch, and $\epsilon$ is a small constant for numerical stabilization.

One may argue that adding BN requires additional hardware resource since Eq.~(\ref{eq:bn}) includes a division and multiplication. However, remembering that BNN requires only the sign of activations, the BN layer at inference time can be realized simply by shifting bias term:
\begin{eqnarray}
    Sign\left\{ BN(x) \right\} &= \begin{cases}
    +1 & x \geq \lfloor \mu - \frac{ \sigma }{\gamma}\beta \rfloor,\\
    -1 & otherwise,
    \end{cases}
\end{eqnarray}
where $\lfloor \cdot \rfloor$ is the floor function.

Owing to the straight-through estimator and BN layers, BNNs can achieve comparable performance with {\ohara{the}} full-precision counterpart. However, we find that naive replacement of DCAE with BNNs results in corruption of extracted image features, which motivates us to develop ``partiall'' binarization technique.

\section{Proposed system}
\label{sec:proposal}
\begin{figure*}[tbp]
  \centering
  \includegraphics[width=17cm]{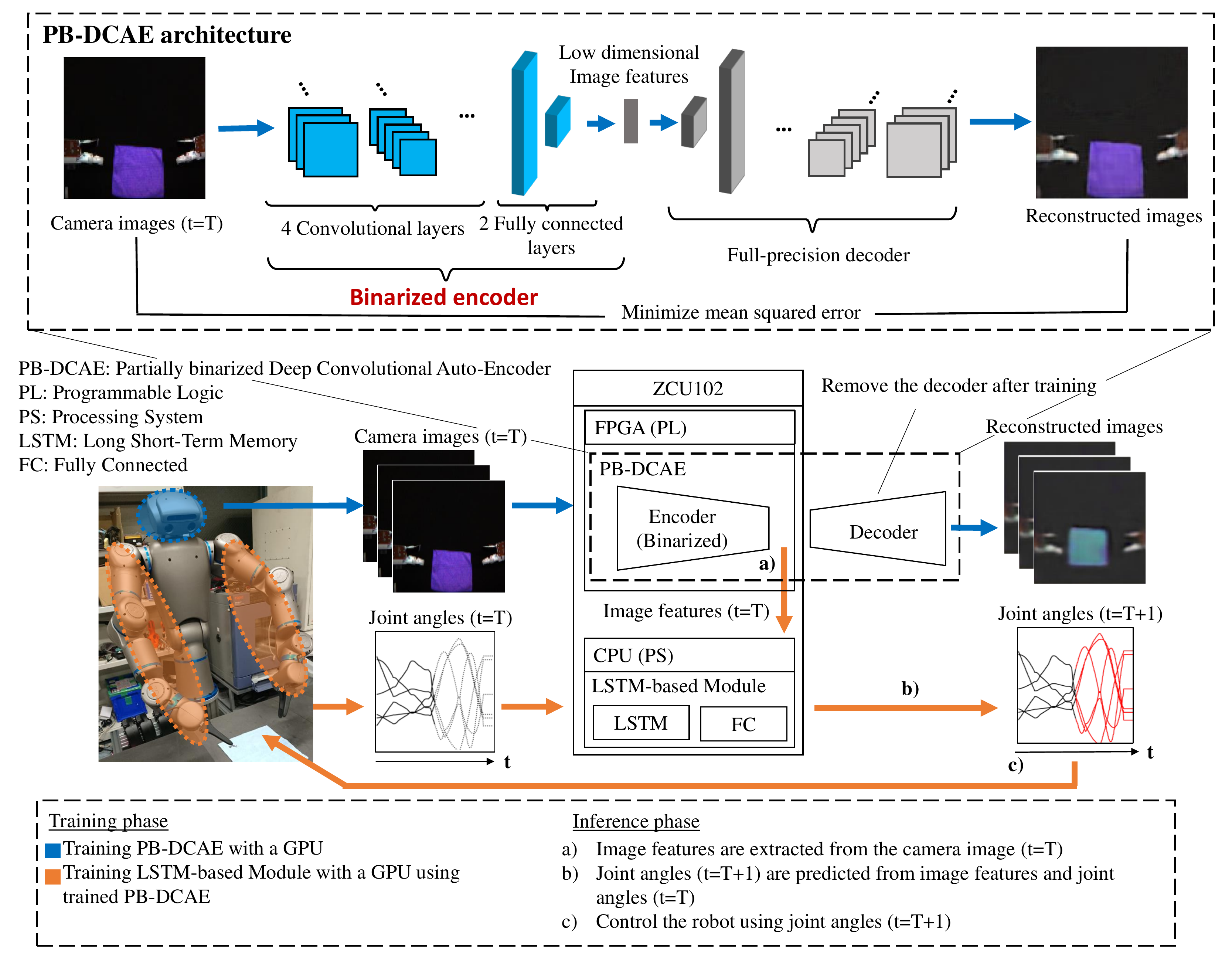}
  \caption{Overall system.}
  \label{fig:system}
\end{figure*}


\subsection{Overview of our system}
Fig.~\ref{fig:system} shows the overall system{\awano{, which is composed of two neural networks: a PB-DCAE to convert raw images into low-dimensional feature vectors and a long-short term memory (LSTM) to generate joint angles of robot arms. Since the LSTM accounts for only {\ohara{1.2}}\% of whole model size, we leave the LSTM unquantized to maintain its performance.
Zynq SoC platform is selected to cope with heterogeneous arithmetic precisions, 
i.e., the PB-DCAE requires binary operations whereas the LSTM requires floating point operations, and
the PB-DCAE and the LSTM are implemented onto a programmable-logic (PL) and processing-system (PS) having an ARM processor, respectively.
Our model is firstly trained by using GPU. Then, the parameters of PB-DCAE are converted to C++ header file and converted to ``.bit'' file via Vivado toolchain to configure the PL. The LSTM is re-implemented by using Numpy from scratch so that it can be efficiently executed on the resource limited ARM processor.}}



\subsection{{\awano{Model architecture}}}
\label{sec:model}
\subsubsection{{\awano{PB-DCAE for image feature extraction}}}
{\awano{To implement the model onto FPGA, the model size should be drastically reduced so that it can fit onto on-chip memory. BNN is a promising technique to compress the model. However, as we demonstrate in Sec.~\ref{sec:experiment}, the naive binarization leads to corruption of extracted image features. To alleviate this problem, we propose ``partially'' binarized DCAE (PB-DCAE), where only the weights and activations in the encoder part are binarized. Since the decoder part is left unchanged, the errors at the output can well backpropagate through the network resulting in the better generalization capability. Note again that the decoder is required only during the training phase and that it is removed for the deployment phase. Hence, PB-DCAE does not incur any additional memory cost compared to fully binarized DCAE.}}


\subsubsection{{\awano{LSTM-based}} Motion generation module}
{\awano{For an RNN part, we employ LSTM which is an extention of naive RNN so that it can take long-term dependencies in sequential data.}} LSTMs are widely used for their expressive power. For example, R. Rahmatizadeh et al. employed a low-cost robotic arm to generate multiple tasks using LSTM\cite{rahmatizadeh2018vision}, and S. Funabashi et al. achieved in-hand object manipulations via CNN-LSTMs\cite{funabashivariable}. The input of LSTM is the robot's joint angles including the gripper commands and image features extracted by PB-DCAE, and outputs the joint angles at the next time step. By recursively feeding the predicted outputs into the input of LSTM, it can make multi-step prediction, which is used to evaluate the performance of the trained LSTM.

\subsection{Training}
\label{sec:construction}


For the ease of stabilization of training, we employed a two-stage training, where the PB-DCAE is trained firstly to yield well self-organized image features, followed by training of LSTM.

{\bf PB-DCAE training:} PB-DCAE is trained to reconstruct input images provided to the input of {\ohara{the}} encoder. Since PB-DCAE has a hourglass-shaped structure, i.e., middle layers have fewer neurons than input or output layers, the network is forced to ``reconstruct'' images by using only degenerated information. Hence, the encoder part tries to find better compressive expressions of input images for the decoder to successfully recover original images. After the training, the decoder part is removed and the parameters of {\ohara{the}} encoder are {\ohara{frozen}}.

\begin{figure}[tbp]
  \centering
  \includegraphics[width=8.5cm]{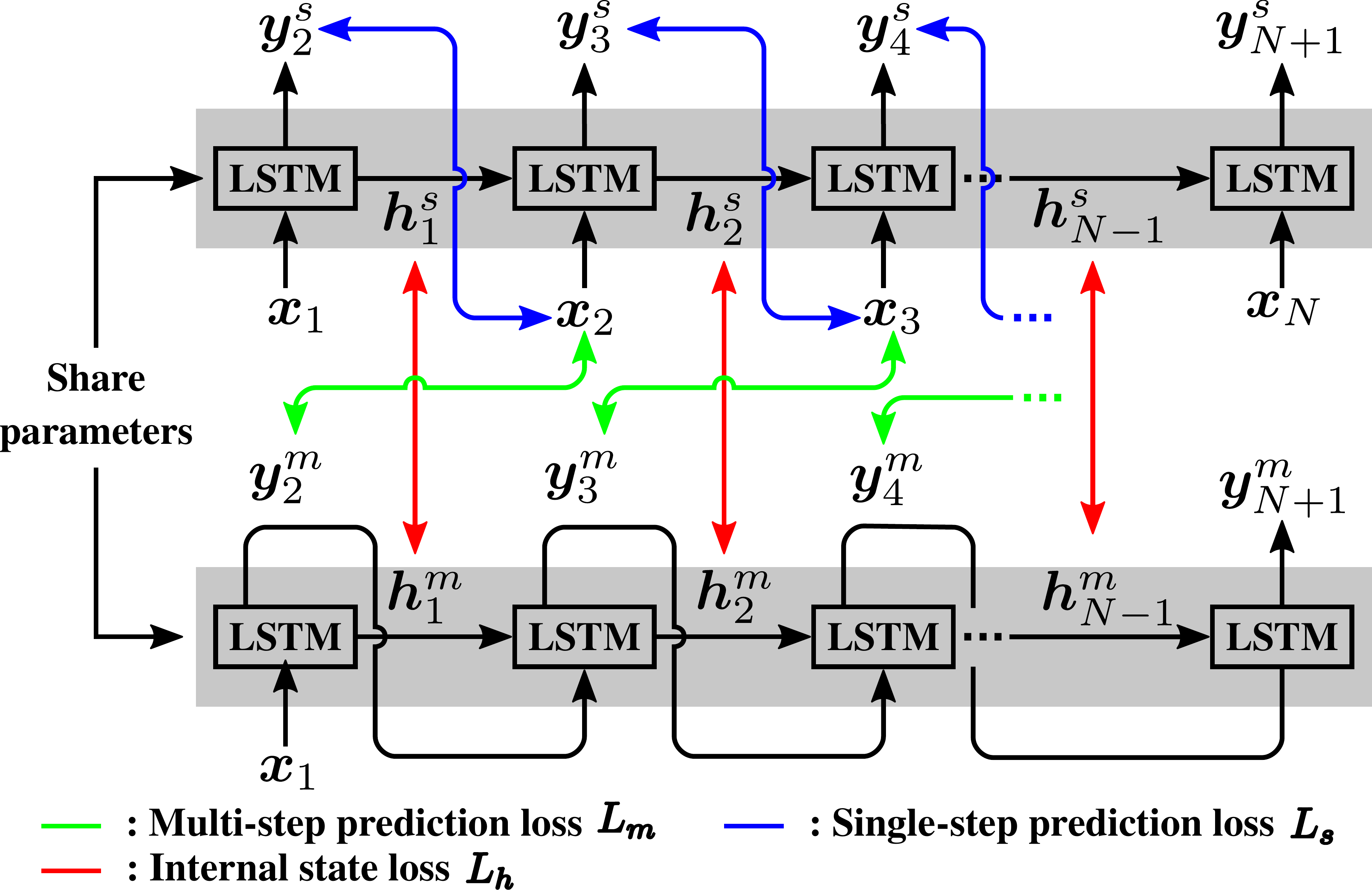}
  \caption{LSTM training setup.}
  \label{fig:lstm}
\end{figure}

{\bf LSTM training:} The image features extracted by the trained PB-DCAE is concatenated with the joint angles and the gripper command to form a visuo-motor sequence, $\bm{X}=(\bm{x}_1,\bm{x}_2,\cdots,\bm{x}_N)$. Here, $\bm{x}_t$ is a vector containing the extracted image feature and the joint angles and the gripper command at time step $t$. $N$ is the sequence length. 
At each time step, the LSTM takes the input $\bm{x}_t$ and the current state $\bm{h}_t$, and outputs $\bm{y}_{t+1}$ and the updated state $\bm{h}_{t+1}$. Let $f_\mathrm{LSTM}(\cdot,\cdot)$ be the mathematical function representing the LSTM's behaviour, the relationship between inputs and outputs can be written as: $\left\{ \bm{y}_{t+1}, \bm{h}_{t+1} \right\} = f_\mathrm{LSTM}( \bm{x}_t, \bm{h}_t )$. By recursively feeding $\left\{ \bm{y}_{t+1}, \bm{h}_{t+1} \right\}$ to the LSTM, it can generate the sequence: ${\bm{Y}}=(\bm{y}_1, \bm{y}_2, \cdots, \bm{y}_t)$. In our application, we want to predict visuo-motor sequence and hence the training objective is to minimize mean-square-loss between $\bm{X}$ and $\bm{Y}$. Unfortunately, however, this ``multi-step'' prediction task is difficult to train and loss tend to diverge frequently especially for the longer sequences.
Hence, to stabilize the training, we propose to exploit single-step prediction loss in combinaiton with multi-step prediction loss. Our training setup is shown in Fig.~\ref{fig:lstm}.

{\bf 1) Multi-step prediction loss:} We firstly compute ``multi-step'' prediction loss by recursively feeding LSTM's outputs to its inputs as shown in the lower part of Fig.~\ref{fig:lstm}. Let $\bm{Y}^\mathrm{m}=\left( \bm{y}_1^\mathrm{m}, \bm{y}_2^\mathrm{m}, \cdots, \bm{y}_t^\mathrm{m} \right)$ be the generated sequence. Then, the multi-step prediction loss $L_m$ is given by:
\begin{align}
L_m &= \frac{1}{2N}\sum_{t=1}^N (\bm{x}_t - \bm{y}_t^m)^2.
\end{align}

{\bf 2) Single-step prediction loss:} Unlike the multi-step prediction setup, the LSTM takes ``teacher'' visuo-motor sequence at each time step, i.e., given an element of visuo-motor sequence $\bm{x}_t$ and the internal state $\bm{h}_t^s$, it computes $\left\{ \bm{y}_{t+1}^s, \bm{h}_{t+1}^s \right\} = f( \bm{y}_t^s, \bm{h}_t^s )$. Then, by recursively feeding only the internal state $\bm{h}_{t+1}^s$, the next output $\bm{y}_{t+2}^s$ is computed as shown in the upper part of Fig.~\ref{fig:lstm}: $\left\{ \bm{y}_{t+2}^s, \bm{h}_{t+2}^s \right\} = f_\mathrm{LSTM} ( \bm{x}_{t+1}, \bm{h}_{t+1}^s )$. Repeating this procedure, we have a sequence of single-step ahead prediction of visuo-motor sequence: $\bm{Y}^s = ( \bm{y}_1^s, \bm{y}_2^s, \cdots, \bm{y}_t^s )$. Then, the single-step prediction loss $L_s$ is given by:
\begin{align}
L_s &= \frac{1}{2N} \sum_{t=1}^N ( \bm{x}_t - \bm{y}_t^s )^2.
\end{align}

  
{\bf 3) Internal state loss:} Since internal states have no teacher signal, the small error may accumulate and cause gradient explosion when making multi-step prediction. Hence, we further introduce an additional loss $L_h$ to make close the internal states for single-step prediction $\bm{h}_t^s$ and those for multi-step prediction $\bm{h}_t^m$:
\begin{align}
L_h &= \frac{1}{2N} \sum_{t=1}^N ( \bm{h}_t^s - \bm{h}_t^m )^2.
\end{align}

 
The final loss function is given by linear combination of the three losses as follows:
\begin{equation}
  L = \alpha L_m + \beta L_s + \gamma L_h, \label{eq:loss}
\end{equation}
where $\alpha$, $\beta$, and $\gamma$ are hyper-parameters.

\subsection{Implementation}
\begin{figure}[tbp]
  \centering
  \includegraphics[width=8.5cm]{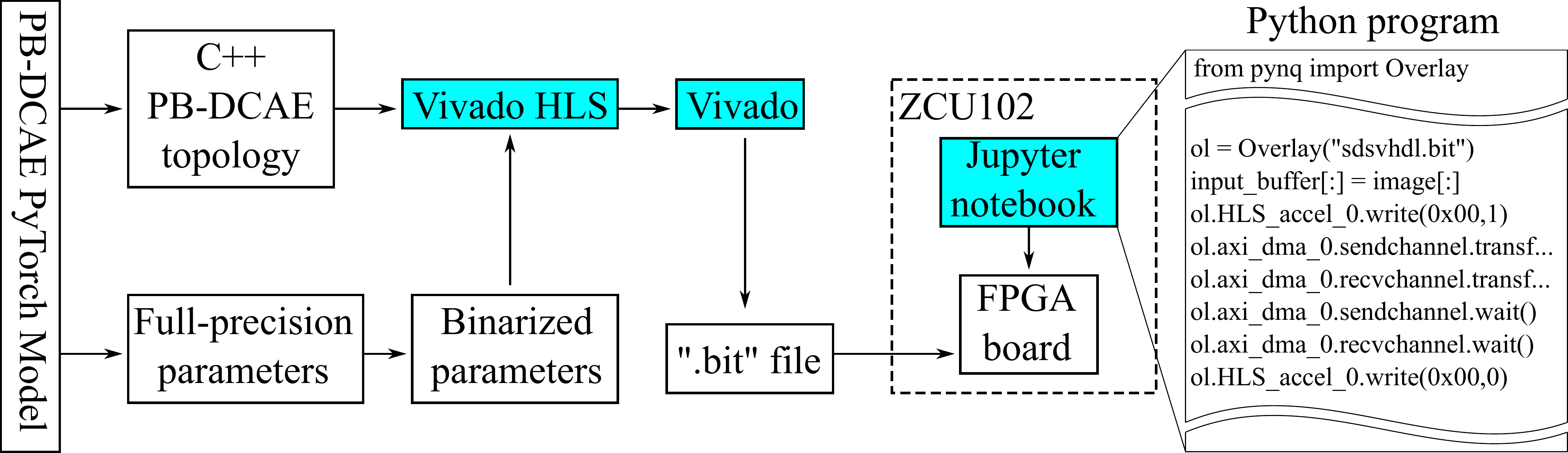}
  \caption{Design flow.}
  \label{fig:design_flow}
\end{figure}

The proposed model
is implemented on Zynq UltraScale+ MPSoC ZCU102 board.
The on-chip RAM and LUT resources of this FPGA board are very limited, so it is necessary to implement a reduced hardware resource usage while maintaining the inference throughput. The overall design flow is shown in Fig.~\ref{fig:design_flow}.
First, PB-DCAE is trained using PyTorch framework and the network parameters are extracted. Next, a C++ model of the proposed PB-DCAE is created and synthesized using two Xilinx Vivado software to yield an executable ``.bit'' file containing the network topology to properly configure the FPGA board. Finally, download all the generated files to a Xilinx ZCU102 FPGA board and test/measure the performance. PYNQ (PYthon Productivity of zynQ) environment is installed onto ZCU102 board so that
we can write programs and view the results via Jupyter Notebook.
{\awano{Note again that}}
The image feature extraction module is implemented on {\awano{PL whereas}} the {\awano{LSTM-based}} motion generation module is implemented on {\awano{PS}}.
Since the {\awano{DCAE-based}}
image feature extraction module accounts for {\ohara{99.94\%}} {\awano{of total multiplications required for single step inference}}
the performance of the entire system {\awano{can be}} greatly improved {\awano{even when the LSTM remains on the PS-side.}}

\begin{figure}[tbp]
  \centering
  \includegraphics[width=8.5cm]{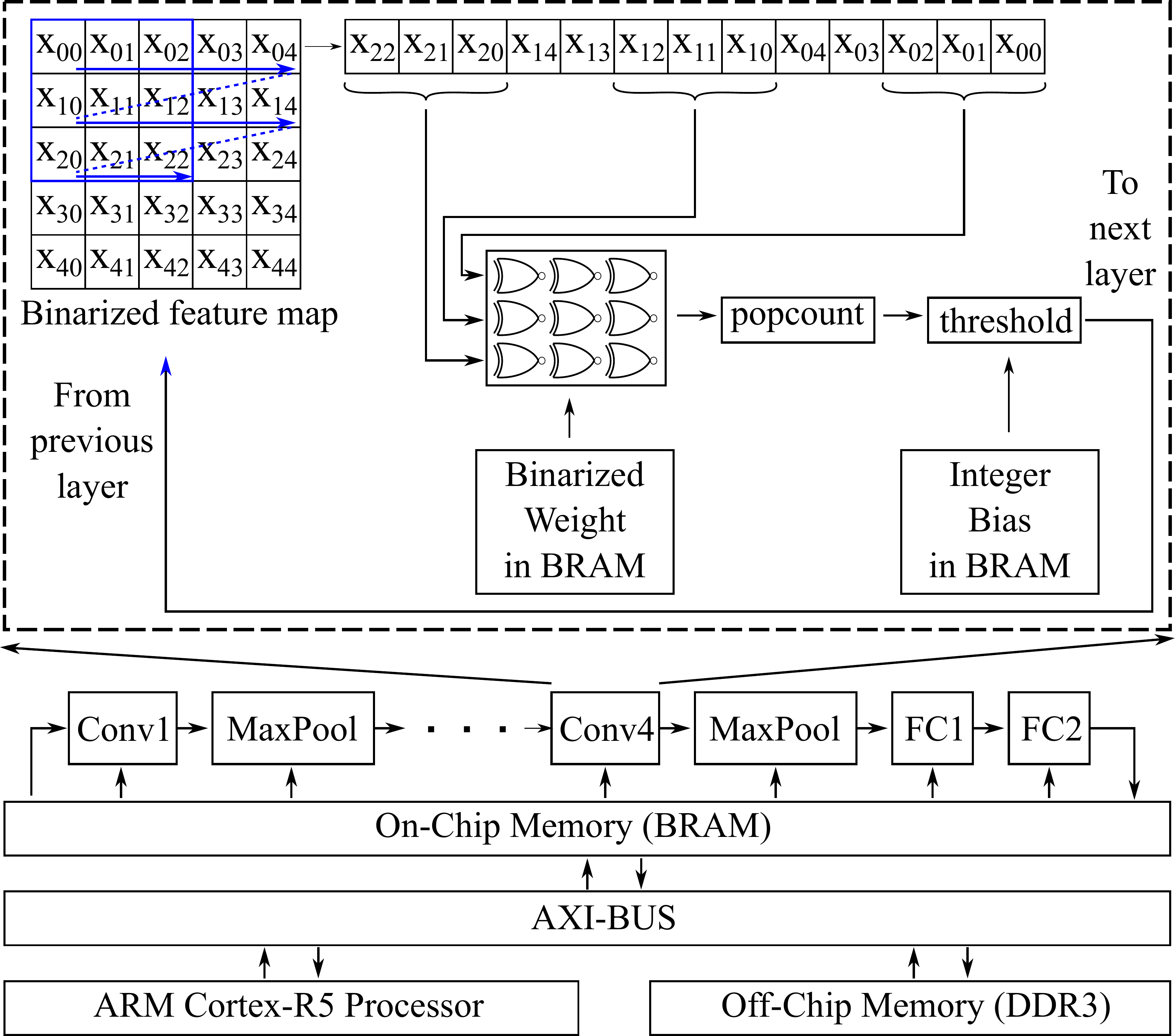}
  \caption{Streaming binarized convolutional circuit}
  \label{fig:fpga_bnn}
\end{figure}

{\bf FPGA implementation:} Fig.~\ref{fig:fpga_bnn} shows the block diagram of PB-DCAE implemented on FPGA. A dedicated computation unit is assigned for each layer to minimize data movements, and each layer is connected via AXI-Stream interface. To process the data stream efficiently with a small hardware resources, shift registers are prepared to hold only the data necessary for the computation and perform convolution.
Binarization allows floating-point multiply-and-accumulate operations to be replaced by XNOR and bit counting, which can be implemented in small circuits.
In addition, although BN requires multipliers and adders, it can be converted to integer comparisons by transforming the expression together with the Sign function that follows, thus reducing the circuit size and implementation.
The model size can be reduced by binarization.
Since the model size is compressed by binarization, all the parameters can be expanded to the on-chip BRAM.
Therefore, feature maps and binarization weights can be stored in on-chip memory for PB-DCAE computation.
Since no external I/O is involved, latency due to communication can be minimized and the system can be executed with low power consumption.

\section{Experiment}
\label{sec:experiment}
\subsection{Experimental setup}
The dataset for the experiments was {\awano{collected by}} using Nextage Open {\awano{developed by Kawada Robotics. It has two arms with grippers and a camera mounted on robot's head. Each arm has six degrees of freedom (DoF). To collect the training data, a robot teleoperating environment is deveoped, where a human operator can control the gripper positions by using 3D mouse. Using this environment, the operator is requested to complete a cloth-folding task, during which camera images (142$\times$142$\times$3=60,492 dimensions, RGB), joint angles of both arms (6$\times$2=12 dimensions), and a gripper command (1 dimension) are recorded. Note here that since we only used right gripper to pick up the cloth, we only recorded the gripper command attached at the right arm.}}
A total of 75 task sequences were collected, 50 of which were used for training and 25 for evaluation. {\awano{As a result, approximately {\ohara{22k}} and {\ohara{11k}} steps of data are used to train and to validate the model, respectively. $\alpha$, $\beta$, and $\gamma$ in Eq.~(\ref{eq:loss}) are set to be 0.1, 1.0, and 0.1, respectively. The model architecture is summarized in Tab.~\ref{tab:param}.}}

\subsection{Experimental result}

\begin{figure}[tbp]
  \centering
  \includegraphics[width=8.75cm]{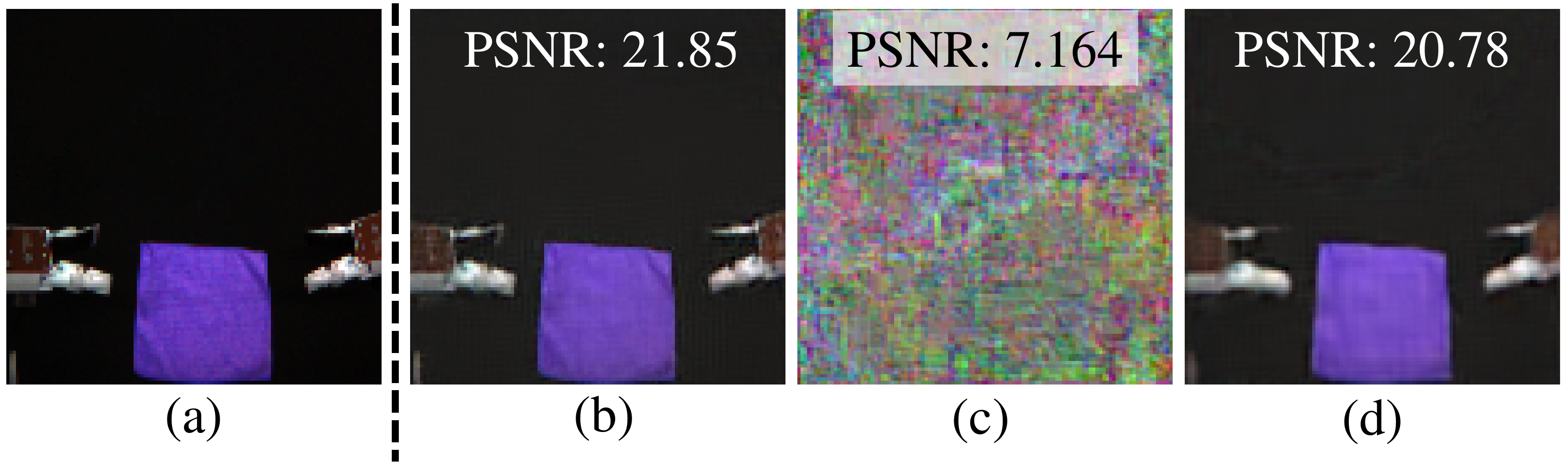}
  \caption{(a) Original and reconstructed images by (b) full-precision DCAE, (c) fully-binarized DCAE, and (d) PB-DCAE.}
  \label{fig:result_DCAE}
\end{figure}

{\awano{To validate the impact of partial binarization, we firstly compare the quality of reconstructed image using three different DCAEs, i.e., a full-precision DCAE, a fully-binarized DCAE, and a PB-DCAE, as shown in Fig.~\ref{fig:result_DCAE}. The left most image is the raw input image which is captured when the robot moves its grippers to pick the towel placed at the center. The remaining three images are those reconstructed by three different DCAEs.

Observing Fig.~\ref{fig:result_DCAE}, we see no noticeable difference between images reconstructed by the full-precision DCAE and the PB-DCAE. However, the fully-binarized DCAE outputs the corrupted image. To quantitatively compare the quality of reconstructed images, peak signal-to-noise ratios (PSNRs), one of the common method to measure the image quality of lossy compression codecs, are calculated. PSNR is expressed by
\begin{eqnarray}
  PSNR = 10 \log_{10} {\frac{MAX_I^2}{MSE}}.
\end{eqnarray}
Higher PSNR means higher image quality and, according to \cite{thomos2005optimized}, PSNR of over 20~dB is sufficient quality for practical applications. As shown in Fig.~\ref{fig:result_DCAE}, the average PSNR for full-precision DCAE was 21.85~dB while that for fully-binarized DCAE was 7.164~dB, which again demonstrates that naive binarization leads to corruption of reconstructed images. We also confirmed that PSNR for PB-DCAE was 20.78~dB which is comparable to that of the full-precision DCAE.

We then examine the impact of binarization on the prediction accuracy of the joint angles and the gripper command. For this purpose, we compare the multi-step prediction accuracy of the LSTM, i.e., given only image features, joint angles, a gripper command at time step $t=1$, the LSTM is requested to predict the whole remaining sequence for $t=2,3,\cdots,N$ by recursively feeding the predicted output into the LSTM. The reconstruction errors calculated by the mean square error between the original and the predicted sequences are 2.98$^\circ$, 6.32$^\circ$, and 3.43$^\circ$ each corresponds to the image features extracted by full-precision DCAE, fully-binarized DCAE, and PB-DCAE, respectively. We again confirmed that the PB-DCAE achieved comparable performance with the full-precision DCAE
{\ohara{although the model size was dramatically reduced from 51.3MB to 2.20MB, as shown Tab.~\ref{tab:param}.}}

We finally compare the energy efficiency of the proposed method with the conventional method relying on a GPU-based platform.
The power consumptions of Zynq ZCU102 board, CPU, and GPU are measured by using Maxim powertool, ``s-tui'' command, and ``nvidia-smi'' command, respectively.
Tab.~\ref{tab:performance} summarizes the processing time, power consumption, and energy efficiency for CPU, GPU, and Zynq platforms. 
Studying Tab.~\ref{tab:performance}, we can confirm that the Zynq platform achieved the highest energy efficiency, which corresponds to 10$\times$ and 3.7$\times$ improvements compared to CPU and GPU platform, respectively, while satisfying the processing speed above 30~FPS which is the upper bound determined by the frame rate of a commercial RGB camera.
}}

\begin{table}[tbp]
  \centering
    \caption{Network architecture and model size.}
    \begin{tabular}{c|c|c|c|c}\hline
      Layer & Out. & In. &Float-besed & Proposed\\
      & F size & F maps & model & model  \\ \hline \hline
      Conv1 & 142$\times$142 &3 & 3.38 KB&  0.105 KB\\
      Max Pool& 70$\times$70& 32& -- & --\\
      Conv2 & 70$\times$70& 32& 72.0 KB & 2.25 KB\\
      Max Pool & 34$\times$34& 64& -- & -- \\
      Conv3 & 34$\times$34& 64& 288 KB & 9.00 KB\\
      Max Pool & 16$\times$16& 128& -- & -- \\
      Conv4 & 16$\times$16& 128& 1.15 MB  &36.0 KB\\
      Max Pool & 7$\times$7& 256& -- & --\\  
      FC1 &1024 &12544 & 50.2 MB & 1.57 MB\\
      FC2 &64 &1024 &256 KB& 8.00 KB\\ \hline
      LSTM1 &100 &77 &280 KB & 280 KB\\
      LSTM2 &100 &100 & 316 KB & 316 KB\\
      FC3 &77 &100 &30.4 KB & 30.4 KB\\ \hline
      Total & -- & -- & 51.3 MB & 2.20 MB\\ \hline
    \end{tabular}
    \label{tab:param}
\end{table}

\begin{table}[tbp]
  \centering
    \caption{Performance comparison.}
    \begin{tabular}{c|c|c|c}\hline
    &   Core i7 6700K & RTX 2080Ti & ZCU102 \\ \hline \hline
    Preprocessing [msec] & 0.759 & 1.47 & 4.45 \\
    Conv1-FC2 [msec] & 12.4 & 2.24 & 15.4\\ 
    LSTM1-FC3 [msec] & 0.420 & 0.315  & 4.48\\
    Total [msec] & 13.7 & 4.03 & 24.3\\ \hline 
    FPS [sec-1] & 73.0 & 248 & 41.1\\ \hline
    Power [W] & 55.4 & 69 & 3.1\\ \hline
    Efficiency [FPS/W] & 1.32 & 3.59 & 13.3\\ \hline
    \end{tabular}
    \label{tab:performance}
\end{table}

\section{Conclusion}
\label{sec:conclusion}
We proposed a lightweight neural network {\awano{named PB-DCAE}}
for {\awano{robotic}} flexible object manipulation.
{\awano{By binarizing only the encoder part of DCAE, we successfully reduced the model size while preserving the performance of the full-precision DCAE.}}
{\awano{PB-DCAE and other peripheral modules are implemented on Zynq ZCU102 board, which demonstrated}}
the model size can be reduced by 95.7\% compared {\awano{to}} {\awano{the full-precision counterpart}}, while the degradation of the predictive accuracy of joint angles is kept to 0.45$^\circ$.
{\awano{Further, we examined the energy efficiency of our system by comparing the same model implemented on Core i7 6700K and RTX 2080 Ti, which demonstrated that 10$\times$ and 3.7$\times$ more efficiency compared the CPU and GPU-based platform, respectively.}}

\bibliographystyle{ieeetr}
\bibliography{IEEEabrv,refs}

\end{document}